\documentclass{article}

\usepackage{microtype}
\usepackage{graphicx}
\usepackage{subcaption}
\usepackage{booktabs} 
\usepackage{float} 
\usepackage[section]{placeins}

\usepackage{hyperref}

\usepackage[preprint]{icml2026}

\usepackage{amsmath}
\usepackage[most]{tcolorbox}
\usepackage{amssymb}
\usepackage{mathtools}
\usepackage{amsthm}

\usepackage[capitalize,noabbrev]{cleveref}

\theoremstyle{plain}

\theoremstyle{definition}

\theoremstyle{remark}

\usepackage[textsize=tiny]{todonotes}

\icmltitlerunning{Decomposing Neural Network Error via Flow-Based Oracles}

\begin{document}

\twocolumn[
  \icmltitle{Beyond the Loss Curve: Scaling Laws, Active Learning, and the Limits of Learning from Exact Posteriors}



  \icmlsetsymbol{equal}{*}

  \begin{icmlauthorlist}
    \icmlauthor{Arian Khorasani}{mila}
    \icmlauthor{Nathaniel Chen}{equal,princton}
    \icmlauthor{Yug D Oswal}{equal,india}
    \icmlauthor{Akshat Santhana Gopalan}{highschool}
    \icmlauthor{Egemen Kolemen}{princton}
    \icmlauthor{Ravid Shwartz-Ziv}{NYU}
  \end{icmlauthorlist}

  \icmlaffiliation{mila}{Mila-Quebec AI Institute, Canada}
  \icmlaffiliation{india}{School of Computer Science and Engineering, Vellore Institute of Technology, India}
  \icmlaffiliation{princton}{Plasma Physics Lab, Princeton University, USA}
  \icmlaffiliation{highschool}{High School Student, John P. Stevens High School, USA}
  \icmlaffiliation{NYU}{Center of Data Science, New York University, New York, USA}

  \icmlcorrespondingauthor{Arian Khorasani}{Arian.Khorasani@mila.quebec}

  \icmlkeywords{Machine Learning, ICML}

  \vskip 0.3in
]


\printAffiliationsAndNotice{\icmlEqualContribution}  

\begin{abstract} How close are neural networks to the best they could possibly do? Standard benchmarks cannot answer this because they lack access to the true posterior $p(y|x)$. We use class-conditional normalizing flows as oracles that make exact posteriors tractable on realistic images (AFHQ, ImageNet). This enables five lines of investigation. \textbf{Scaling laws:} Prediction error decomposes into irreducible aleatoric uncertainty and reducible epistemic error; the epistemic component follows a power law in dataset size, continuing to shrink even when total loss plateaus. \textbf{Limits of learning:} The aleatoric floor is exactly measurable, and architectures differ markedly in how they approach it: ResNets exhibit clean power-law scaling while Vision Transformers stall in low-data regimes. \textbf{Soft labels:} Oracle posteriors contain learnable structure beyond class labels: training with exact posteriors outperforms hard labels and yields near-perfect calibration. \textbf{Distribution shift:} The oracle computes exact KL divergence of controlled perturbations, revealing that shift \emph{type} matters more than shift magnitude: class imbalance barely affects accuracy at divergence values where input noise causes catastrophic degradation. \textbf{Active learning:} Exact epistemic uncertainty distinguishes genuinely informative samples from inherently ambiguous ones, improving sample efficiency. Our framework reveals that standard metrics hide ongoing learning, mask architectural differences, and cannot diagnose the nature of distribution shift.
\end{abstract}


\section{Introduction}

Every paper reporting test accuracy implicitly asks: how good is this model? But good compared to what? Without access to the true posterior $p(y|x)$, we cannot tell whether a model is at 50\% or 99\% of the theoretical maximum. When a loss curve flattens, we cannot distinguish irreducible noise (aleatoric uncertainty) from gaps in the model's knowledge (epistemic uncertainty)~\citep{gal_dropout_2016,guo_calibration_2017}. Neural Scaling laws~\citep{kaplan_scaling_2020} tell us loss decreases as $N^{-\alpha}$, but loss conflates both \cite{hoffmann_training_2022}. When performance degrades under distribution shift, we cannot tell whether the shift was large or small~\citep{hendrycks_benchmarking_2019} \cite{gal_dropout_2016}.

A class-conditional normalizing flow~\citep{dinh_density_2017,kingma_glow_2018} trained on real images provides a concrete reference point. The flow defines an explicit $p_\theta(x\mid y)$ from which we can both sample and evaluate exact likelihoods. We treat this not as an approximation to nature, but as the complete specification of a synthetic world in which Bayes-optimal posteriors are computable in closed form~\citep{dinh_density_2017}. In this world, the expected cross-entropy of any classifier decomposes exactly:
\begin{equation*}
    \mathcal{L}(q_\theta) = \underbrace{\mathbb{E}_x\!\left[H(p(y|x))\right]}_{\text{aleatoric (irreducible)}} + \underbrace{\mathbb{E}_x\!\left[\mathrm{KL}(p(y|x) \| q_\theta(y|x))\right]}_{\text{epistemic (reducible)}}
\end{equation*}

\begin{figure*}[t]
    \centering
    \includegraphics[width=\textwidth]{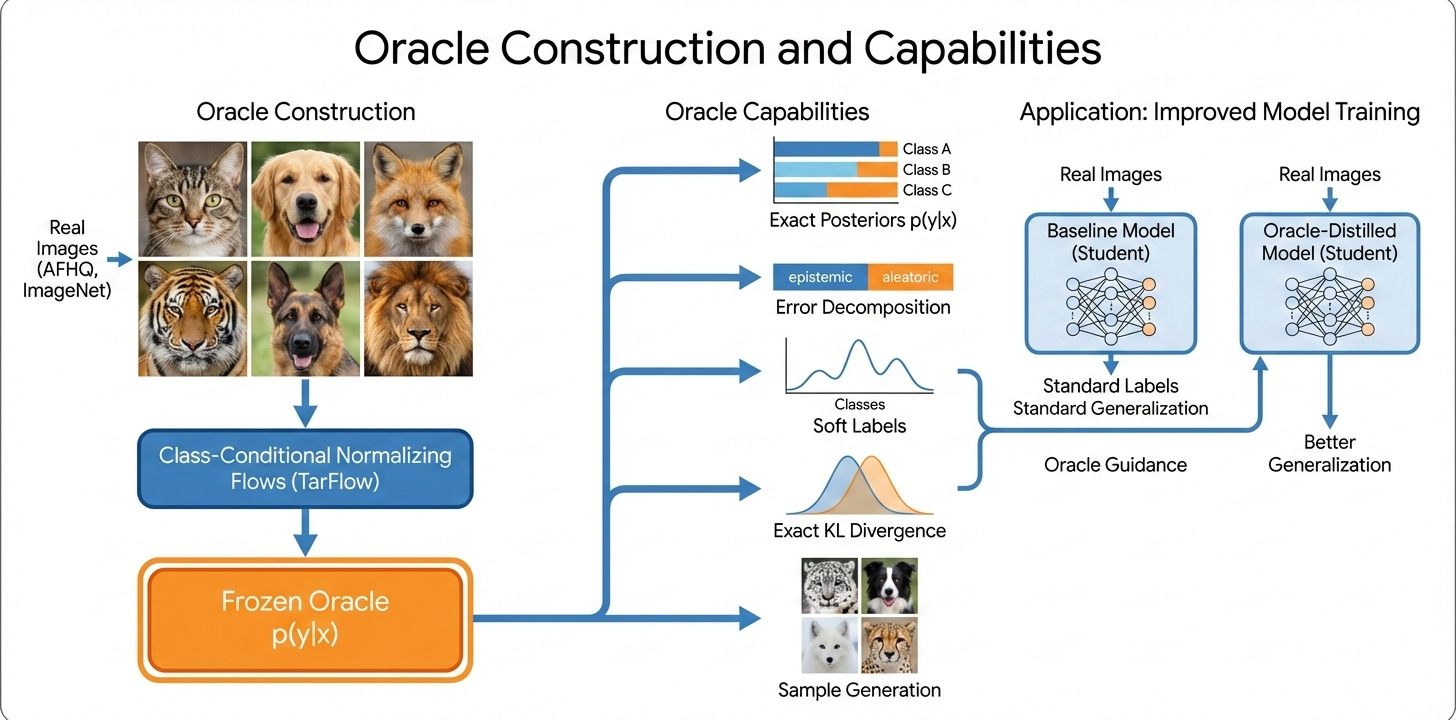}
    \caption{\textbf{Normalizing flows enable exact posterior computation on realistic images.}
\textbf{(Left)}~We train class-conditional flows on AFHQ/ImageNet, creating a frozen oracle with tractable likelihoods.
\textbf{(Center)}~The oracle decomposes prediction error into aleatoric (irreducible) and epistemic (reducible) components, quantities hidden in standard benchmarks.
\textbf{(Right)}~We apply this framework to diagnose scaling laws, quantify distribution shift with exact KL divergence, and train models with exact soft labels.}
    \label{fig:oracle_overview}
\end{figure*}

This decomposition, intractable on any standard benchmark, is a direct consequence of Bayes-optimal risk decomposition and is the basis for all our experiments. As illustrated in Figure~\ref{fig:oracle_overview}, we train state-of-the-art normalizing flows on realistic image datasets and conduct a thorough validation to confirm the generated images match real data statistics and are not memorized (Section~\ref{sec:validation}). We find:

\begin{itemize}
    \item \textbf{Scaling laws beyond the loss curve.} Epistemic error follows a power law $\mathrm{KL} \propto N^{-\alpha}$ that continues to shrink even when total loss plateaus. The exponents differ across architectures.
    \item \textbf{The limits of learning are exactly measurable.} The aleatoric floor is a hard limit no architecture can beat. Architectures differ markedly in how they approach it: ResNets show clean power-law scaling while Vision Transformers stall without pretraining.
    \item \textbf{Soft labels from exact posteriors enable better learning.} Training with $p(y|x)$ instead of argmax labels outperforms hard labels and yields near-perfect calibration~\cite{szegedy_rethinking_2015}.
    \item \textbf{Distribution shift is exactly quantifiable.} The oracle computes $\mathrm{KL}[p_{\text{shifted}} \| p_{\text{baseline}}]$ exactly, revealing that shift \emph{type} matters far more than magnitude: class imbalance barely affects accuracy at divergence values where input noise causes collapse.
    \item \textbf{Active learning with ground-truth uncertainty.} Exact epistemic uncertainty distinguishes genuinely informative samples from inherently ambiguous ones, improving sample efficiency over standard acquisition functions that confound aleatoric and epistemic uncertainty~\cite{gal_deep_2017}.
\end{itemize}

Our oracle defines a synthetic world, not nature. But the phenomena it reveals (hidden epistemic progress, architectural scaling gaps, misleading divergence measures) are properties of the \emph{models and metrics}, not the data source \cite{belghazi_mine_2021}.

\section{The Oracle Framework}

We build an oracle benchmark by training class-conditional normalizing flows that define tractable densities $p_\theta(x\mid y)$ on high-dimensional inputs~\cite{papamakarios_normalizing_2021}. Once the flows are trained and frozen, they specify a synthetic world in which the posterior is available by Bayes' rule,
\begin{equation}
\label{eq:oracle-posterior}
    p(y\mid x) = \frac{p_\theta(x\mid y)\,\pi_y}{\sum_{y'} p_\theta(x\mid y')\,\pi_{y'}}.
\end{equation}
Because likelihoods are tractable, we can compute Bayes-optimal posteriors exactly under the oracle model (up to floating-point arithmetic) via Bayes' rule~\citep{dinh_density_2017}. This in turn makes our loss decomposition and controlled scaling, shift, and active-learning experiments possible. Practical details on training and scaling to many classes (ImageNet) are provided in the appendix \cite{deng_imagenet_2009}.

\subsection{Flows as Oracles}

A normalizing flow defines an invertible mapping between inputs $x$ and latent variables $z$~\citep{dinh_density_2017,kingma_glow_2018}: the forward map $f_\theta$ sends $x \mapsto z$, while the inverse $f_\theta^{-1}$ generates samples by mapping $z \mapsto x$ for $z \sim \mathcal{N}(0,I)$. The key property is that densities are tractable \cite{papamakarios_normalizing_2021}:
\begin{equation}
    \log p_\theta(x) = \log p_Z(f_\theta(x)) + \log \left| \det J_{f_\theta}(x) \right|.
\end{equation}
We train one flow per class, yielding class-conditional densities $p_\theta(x\mid y)$, and then freeze the parameters. For any input $x$, we compute the oracle posterior via Bayes' rule (Eq.~\ref{eq:oracle-posterior}).

Unlike typical Bayesian methods that rely on sampling or variational approximations, this computation is exact \emph{under the oracle model}~\cite{alemi_deep_2019} since the likelihoods and class prior are explicitly known: given the trained flows and the chosen class prior $\pi$, $p(y\mid x)$ is computed in closed form (implemented in log space with log-sum-exp normalization for numerical stability). The oracle therefore returns the true posterior for the synthetic world defined by the flow.  

\subsection{What Does ``Truth'' Mean Here?}                                                     
  Here, ``true'' means Bayes-optimal with respect to the oracle world induced by the trained flows. The oracle world is defined by (i) the per-class flow densities $p_\theta(x\mid y)$, (ii) the exact preprocessing pipeline used during flow training and sampling, and (iii) a class prior $\pi_y$. All ``exact'' statements are with respect to this specification: for the oracle world, we can compute the Bayes posterior $p(y\mid x)$ (Eq.~\ref{eq:oracle-posterior}) and therefore quantities such as the aleatoric term $\mathbb{E}[H(p(y\mid x))]$ and epistemic term $\mathbb{E}[\mathrm{KL}(p\,\|\,q_\theta)]$ exactly under the oracle model (Section~\ref{sec:decomposition}).

  Unless otherwise stated, we use a uniform prior over classes ($\pi_y = 1/K$). In the class-imbalance shift experiments, we modify the class prior $\pi$ while keeping $p_\theta(x\mid y)$ fixed \cite{chawla_smote_2002}, isolating prior shift from conditional shift.

  The main limitation is external validity: conclusions are only useful insofar as samples from the oracle world match the statistical and semantic properties of the real data used to fit the flows.

\subsection{Oracle Validation}
\label{sec:validation}

\begin{tcolorbox}[colback=blue!5, colframe=blue!40, boxrule=0.5pt, left=4pt, right=4pt, top=2pt, bottom=2pt]
\textbf{Bottom line:} Generated images are statistically similar to the original data across six metrics, and classifiers trained on oracle samples approach the Bayes-optimal bound~\cite{salimans_improved_2016}.
\end{tcolorbox}

Because the oracle provides exact posteriors, we can compute the Bayes-optimal accuracy for the oracle world: 99.77\% on AFHQ~\cite{choi_stargan_2020} (Bayes error 0.23\%). If classifiers trained on oracle data approach this bound, the oracle is producing learnable, well-structured samples. We trained five architectures on 50K oracle samples. ConvNeXt~\cite{liu_convnet_2022} reaches 98.0\%, ResNet~\cite{he_deep_2015} 97.7\%, Swin~\cite{liu_swin_2021} 97.7\%, MobileNet~\cite{howard_searching_2019} 97.7\%, and ViT-Base~\cite{dosovitskiy_image_2021} 97.0\% (Table~\ref{tab:self-validation}). The remaining 1.8--2.8\% gap closes with longer training. On ImageNet-64~\cite{chrabaszcz_downsampled_2017}, the Bayes error is 4.031\%.

We validated distributional quality using six metrics. \textbf{FID}~\cite{heusel_gans_2018} measures feature distribution distance (28.44 on AFHQ, 13.48 on ImageNet-64), higher than diffusion models because flows trade sharpness for exact likelihoods. \textbf{Inception Score}~\cite{salimans_improved_2016} evaluates quality and diversity ($6.24 \pm 3.57$ on AFHQ, $32.57 \pm 4.47$ on ImageNet-64). \textbf{Manifold coverage} confirms broad support (90\% on AFHQ, 89.9\% on ImageNet-64). \textbf{Feature variance match} shows semantic diversity is preserved (83--92\% on AFHQ, 72--93\% on ImageNet-64). \textbf{Memorization check}: only 36\% of AFHQ samples (6.5\% on ImageNet-64) have a training neighbor within feature distance 10, and visual inspection confirms these share pose/lighting but depict distinct individuals. Full metrics and protocol in Appendix~\ref{app:validation}, Table~\ref{tab:oracle-quality}.

\begin{table}[t]
\centering
\caption{\textbf{Classifiers approach the Bayes-optimal bound on oracle data.} Self-validation on AFHQ: all architectures reach 97--98\% accuracy, within 2--3\% of the theoretical optimum (99.8\%).}
\label{tab:self-validation}
\begin{tabular}{lccc}
\toprule
Architecture & Accuracy (\%) & Error (\%) & Gap (\%) \\
\midrule
ConvNeXt & 98.03 & 1.97 & 1.75 \\
ResNet-18 & 97.73 & 2.27 & 2.05 \\
Swin & 97.68 & 2.32 & 2.10 \\
MobileNet & 97.65 & 2.35 & 2.12 \\
ViT-Base & 96.98 & 3.02 & 2.80 \\
\bottomrule
\end{tabular}
\end{table}

\subsection{Implementation}

We use TarFlow~\citep{zhai_normalizing_2025}, a recent normalizing flow that achieves strong likelihood scores on AFHQ and ImageNet. Training follows standard practice: dequantization, logit preprocessing, and maximum likelihood optimization. We train separate flows per class on AFHQ (3 classes: cat, dog, wild) and ImageNet \cite{deng_imagenet_2009} (up to 1000 classes at 64$\times$64 and 128$\times$128 resolution). After training, we freeze the flows and generate 50,000 labeled samples with oracle posteriors. We compute posteriors in log space and normalize with log-sum-exp for numerical stability. Full architecture and systems details (including throughput and scaling across many classes) appear in the appendix.

\section{Decomposing Prediction Error}
\label{sec:decomposition}

Why does this decomposition matter? In practice, a flattening loss curve is ambiguous: it could mean the model has learned everything learnable, or it could mean the model is stuck while reducible error remains. Distinguishing these cases determines whether more data, more capacity, or a different task formulation is needed. Similarly, calibration methods aim to match model confidence to true probabilities, but without access to true posteriors, calibration can only be evaluated against empirical frequencies~\cite{guo_calibration_2017}. Access to the true posterior $p(y|x)$ enables a decomposition that resolves both questions. Consider a classifier $q_\theta(y|x)$ and measure its expected cross-entropy:
\begin{equation}
    \mathcal{L}(q_\theta) = \mathbb{E}_{x}\left[-\sum_y p(y|x) \log q_\theta(y|x)\right]
\end{equation}
This loss splits cleanly into two terms:
\begin{equation}
    \mathcal{L}(q_\theta) = \underbrace{\mathbb{E}_x\left[H(p(y|x))\right]}_{\text{aleatoric}} + \underbrace{\mathbb{E}_x\left[\mathrm{KL}(p(y|x) \| q_\theta(y|x))\right]}_{\text{epistemic}}
\end{equation}

The \textbf{aleatoric} term is the entropy of the true posterior averaged over inputs. It reflects irreducible uncertainty, or the ambiguity inherent in the data that no classifier can resolve. An image of a cat-like dog has high $H(p(y|x))$ regardless of model quality.

The \textbf{epistemic} term measures the gap between the model's beliefs and truth \cite{gal_dropout_2016}. A perfect model achieves $\mathrm{KL}=0$; any deviation indicates something learnable that the model has not yet captured.

On standard benchmarks, we observe only total loss. When it plateaus, we cannot tell whether the model has reached the aleatoric floor or whether epistemic error remains. Our oracle makes both terms measurable. This distinction matters: if epistemic error dominates, more data or capacity should help; if aleatoric error dominates, improvements require changing the task itself.

\section{Experiments}

\paragraph{Setup.} We trained TarFlow~\cite{zhai_normalizing_2025} on AFHQ~\cite{choi_stargan_2020} (3 classes: cat, dog, wild; $\sim$15K images) and ImageNet~\cite{deng_imagenet_2009} at 64$\times$64 resolution. Oracle quality is validated in Section~\ref{sec:validation}. We generated labeled samples with exact posteriors for all experiments. We trained classifiers (ResNet-50~\cite{he_deep_2015}, ViT-Base~\cite{dosovitskiy_image_2021}, ConvNeXt~\cite{liu_convnet_2022}, Swin~\cite{liu_swin_2021}, MobileNetV3~\cite{howard_searching_2019}) using default hyperparameters on varying amounts of oracle data, from $N{=}100$ to $N{=}10{,}000$ (up to $N{=}45,000$ for ImageNet-64) samples. All results are averaged over 3 seeds with standard deviation shown.

\subsection{Scaling Laws Beyond the Loss Curve}
\label{sec:scaling}

\begin{tcolorbox}[colback=blue!5, colframe=blue!40, boxrule=0.5pt, left=4pt, right=4pt, top=2pt, bottom=2pt]
\textbf{Bottom line:} Epistemic error follows a clean power law in dataset size, continuing to shrink even when total loss plateaus. Standard metrics hide this ongoing learning.
\end{tcolorbox}

We trained classifiers on varying amounts of oracle data and decomposed prediction error into aleatoric and epistemic components \cite{kaplan_scaling_2020} (Section~\ref{sec:decomposition}). For each model, we computed total cross-entropy loss (what standard benchmarks measure), the aleatoric component $\mathbb{E}[H(p(y|x))]$, and the epistemic component $\mathbb{E}[\mathrm{KL}(p(y|x) \| q_\theta(y|x))]$.

\begin{figure*}[t]
\centering
\includegraphics[width=0.75\textwidth]{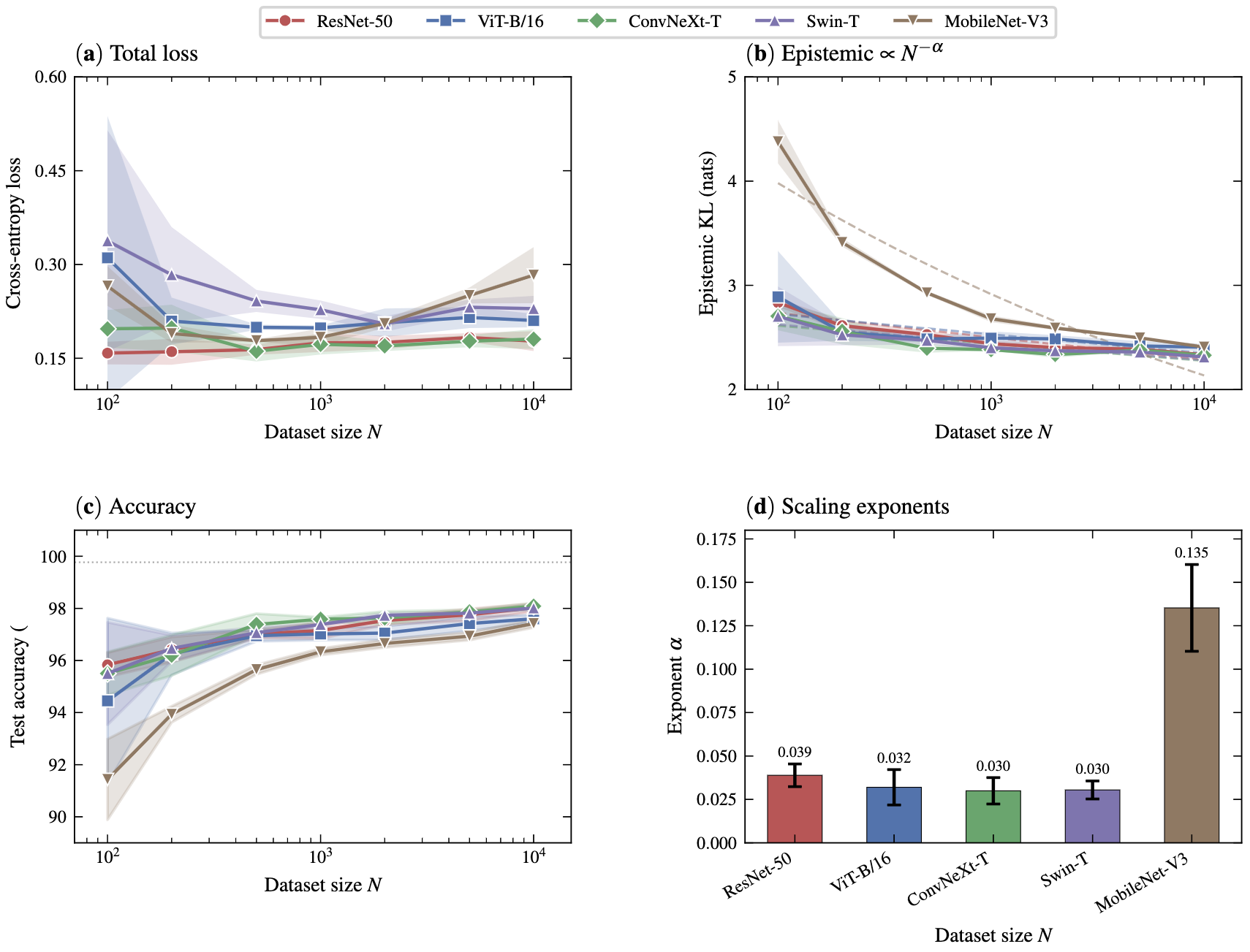}
\caption{\textbf{Epistemic error follows a power law even when total loss plateaus.}
\textbf{(a)}~Total cross-entropy decreases with dataset size; MobileNet shows the highest loss, ResNet/ConvNeXt the lowest.
\textbf{(b)}~Epistemic uncertainty (KL from oracle) follows $N^{-\alpha}$; dashed lines are power-law fits. This decay continues even when total loss appears flat.
\textbf{(c)}~Accuracy improves from 92--96\% at $N{=}100$ toward 97--98\% at $N{=}10{,}000$.
\textbf{(d)}~Scaling exponents reveal architectural differences: MobileNet improves fastest ($\alpha{=}0.135$), ViT stalls without pretraining ($\alpha{=}0.032$).}
\label{fig:scaling}
\end{figure*}

Figure~\ref{fig:scaling} shows the result. Total cross-entropy loss decreases with dataset size on a log-log scale, consistent with~\citet{kaplan_scaling_2020}. MobileNet exhibits the highest loss throughout, ResNet and ConvNeXt the lowest. But the total loss curve obscures what is actually happening.

The epistemic component behaves differently. On a log-log plot, it traces a nearly perfect straight line: $\mathrm{KL} \propto N^{-\alpha}$ (dashed lines in Figure~\ref{fig:scaling}b)~\cite{hoffmann_training_2022}. This decay \emph{continues even when total loss appears to plateau}, because the aleatoric floor dominates the total at large $N$. Standard metrics miss this ongoing learning because they conflate the two error sources.

The scaling exponents $\alpha$ differ across architectures (Figure~\ref{fig:scaling}d): ResNet-50 ($\alpha{=}0.039 \pm 0.007$), ViT-Base ($0.032 \pm 0.010$), ConvNeXt ($0.030 \pm 0.008$), Swin ($0.030 \pm 0.005$), and MobileNetV3 ($0.135 \pm 0.025$). These quantify how efficiently each architecture converts additional data into reduced uncertainty; MobileNet's much higher $\alpha$ indicates faster epistemic decay rate, yet it starts from a higher error floor, suggesting a less efficient initial representation. Calibration error (ECE) also decreases with $N$, but at a different rate than epistemic error, confirming that calibration and posterior approximation quality are distinct phenomena~\cite{guo_calibration_2017}.

\paragraph{The aleatoric floor.} On AFHQ, the aleatoric uncertainty is exactly 0.012 nats (Bayes error 0.23\%; see Section~\ref{sec:validation}), with mutual information $I(X;Y) = 1.577$ bits (normalized MI = 99.5\%). On ImageNet-64, the aleatoric floor is 0.10 nats (Bayes error 4.031\%), higher due to the greater number of classes and inherent inter-class ambiguity.

\paragraph{ImageNet replication.} 
Figure~\ref{fig:scaling_laws_imgnet64} shows scaling laws on ImageNet-64 up to 45K samples, a substantially more challenging dataset than AFHQ due to its fine-grained structure and 1000-way label space. Despite this higher intrinsic entropy, the total cross-entropy loss decreases rapidly with dataset size, at a faster absolute rate than in AFHQ, reinforcing the classical finding that larger datasets continue to yield measurable gains even when the absolute loss remains dominated by intrinsic class complexity.

The aleatoric component remains effectively constant at approximately 0.10 nats across all dataset sizes, confirming that all models converge to the same irreducible uncertainty induced by the oracle conditional distribution. As in AFHQ, this floor is architecture-independent and invariant to dataset size, reflecting the Bayes error of the oracle world rather than model capacity. The higher value relative to AFHQ (0.012 nats) is consistent with ImageNet-64’s greater inter-class overlap and multimodality.

Epistemic uncertainty decays with dataset size and corroborates the power-law behaviour observed on AFHQ. ImageNet-64, however, exhibits a brief plateau in the mid-range (2,000-10,000 samples), reflecting a representation-transition regime in which models do not immediately convert additional data into reduced epistemic uncertainty. Beyond this regime, all architectures resume clear power-law decay, reinforcing that epistemic uncertainty, alongside total loss, is a distinct and reliable indicator of continued learning at scale.

The resulting exponents show both similarities and divergences relative to AFHQ. ResNet-50 exhibits the highest scaling rate ($\alpha{=}0.019 \pm 0.006$), closely followed by ViT-Base ($0.018 \pm 0.004$), mirroring their behaviour on AFHQ. ConvNeXt achieves only a weak decay rate ($0.003 \pm 0.003$, lower than ResNet and ViT-Base as in AFHQ), indicating limited epistemic improvement per additional sample, likely due to architectural biases that mismatch the ImageNet-64 feature geometry. Swin-T displays the strongest positive exponent ($0.027 \pm 0.01$), consistent with hierarchical transformers benefiting disproportionately from larger, more diverse datasets. 

MobileNet-V3 is an exception: its epistemic curve exhibits a pronounced non-monotonicity. We attribute this to MobileNet's compressed architecture and depthwise-separable convolutions, which yield an inefficient initial representation incapable of faithfully capturing ImageNet-64's high inter-class diversity. At small dataset sizes, the model overfits and becomes overconfident, producing an artificially low epistemic floor. As the dataset expands into the mid-scale regime ($\approx$2k--10k samples), this brittle representation is forced to reorganize, triggering a spike in loss, calibration error, and epistemic uncertainty. Once past this regime, MobileNet resumes clean power-law decay, confirming that the underlying epistemic scaling law still holds once an adequate representation is established.

Together with the monotonic rise in test accuracy, these trends show that accuracy, calibration, and epistemic uncertainty respond differently to data scaling, with calibration showing partial signatures of the behaviour made explicit by epistemic uncertainty.

\vspace{-1em}

\paragraph{Architectures differ in how they approach optimality.} Table~\ref{tab:architecture-comparison} shows that ResNet-18 exhibits clean power-law scaling: epistemic error drops from 0.16 to 0.026 as data increases from $N{=}100$ to $N{=}40{,}000$ (a $6\times$ reduction). ViT-Base behaves differently: it starts competitive at $N{=}100$ but then stalls. By $N{=}40{,}000$, ViT has improved by less than $1.2\times$. Both architectures achieve similar total loss at large $N$; the difference is in \emph{how} they approach optimality. ViTs, lacking pretraining, fail to extract geometric structure from small samples \cite{dosovitskiy_image_2021}. This is consistent with~\citet{dosovitskiy_image_2021}, but our framework quantifies the gap in information-theoretic terms.

\begin{table}[t]
\caption{\textbf{ResNets reduce epistemic error $6\times$; ViTs stall without pretraining.} Epistemic KL (nats) vs.\ dataset size on AFHQ. ResNet exhibits clean power-law scaling from 0.16 to 0.026; ViT improves by only $1.2\times$ over the same range.}
\label{tab:architecture-comparison}
\begin{center}
\begin{small}
\begin{tabular}{lcc}
\toprule
Dataset size $N$ & ResNet-18 & ViT-B-16 \\
\midrule
100 & 0.160 & 0.131 \\
1,000 & 0.090 & 0.141 \\
5,000 & 0.058 & 0.099 \\
10,000 & 0.033 & 0.117 \\
40,000 & 0.026 & 0.117 \\
\bottomrule
\end{tabular}
\end{small}
\end{center}
\end{table}

\subsection{Soft Labels Enable Better Learning}
\label{sec:softlabels}

Our oracle can generate exact posterior distributions $p(y|x)$ as training labels, not just the argmax class. Training with these soft labels outperforms hard-label training at 4 out of 5 dataset sizes, with accuracy gains up to ${\sim}$1\%. Models trained on soft labels also achieve near-perfect calibration (ECE $= 0.018$)~\cite{guo_calibration_2017}, confirming the posteriors encode learnable structure beyond class labels~\cite{szegedy_rethinking_2015,hinton_distilling_2015}. Full results in Appendix~\ref{app:softlabels}.

\subsection{Distribution Shift: Exact Quantification via Oracle}
\label{sec:distshift}

\begin{tcolorbox}[colback=blue!5, colframe=blue!40, boxrule=0.5pt, left=4pt, right=4pt, top=2pt, bottom=2pt]
\textbf{Bottom line:} The oracle computes exact KL divergence of controlled perturbations. Class imbalance barely affects accuracy at KL values where input noise causes collapse. Shift type matters far more than shift magnitude.
\end{tcolorbox}

On standard benchmarks, distribution shift is observed but never measured: we see accuracy drop but cannot quantify how much the distribution actually changed. Our oracle computes $\mathrm{KL}[p_{\text{shifted}} \| p_{\text{baseline}}]$ exactly by evaluating the true log-likelihood $\log p(x|y)$ under both distributions. This enables controlled experiments that are impossible on real benchmarks: we introduce perturbations of known type and magnitude, measure the exact divergence, and observe how models degrade.

We introduce two types of controlled perturbation to the oracle distribution:
\begin{itemize}
    \item \textbf{Class imbalance:} Shifting the class prior from uniform to skewed ratios (e.g., 40/35/25 up to 70/20/10), which increases KL divergence through the marginal $p(y)$ \cite{chawla_smote_2002}.
    \item \textbf{Gaussian noise:} Adding noise with standard deviation $\sigma \in \{0.05, 0.10, 0.15\}$ to generated images, which increases KL divergence through the conditional $p(x|y)$.
\end{itemize}

\begin{figure*}[t]
\centering
\includegraphics[width=0.95\textwidth]{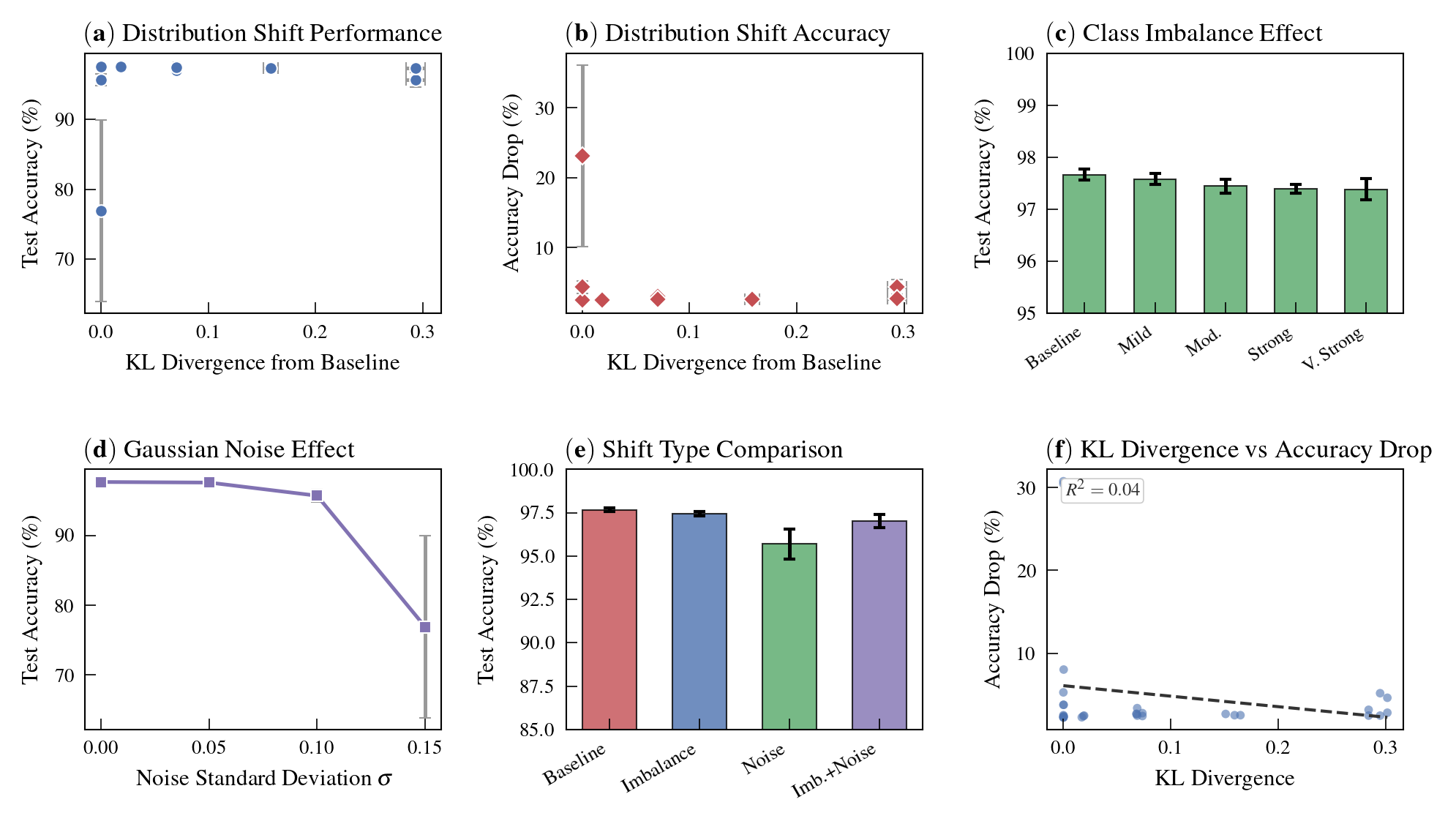}
\caption{\textbf{What shifts matters more than how much: KL magnitude alone poorly predicts performance ($R^2{=}0.04$).}
\textbf{(a)}~Test accuracy vs.\ exact KL divergence: class imbalance (blue) maintains ${\sim}$97\% accuracy; Gaussian noise (low points) collapses to ${\sim}$77\% at comparable KL.
\textbf{(b)}~Accuracy drop vs.\ KL: imbalance (red diamonds) clusters near zero regardless of divergence magnitude.
\textbf{(c)}~Prior shift alone barely affects accuracy (97.4--97.7\% across all imbalance levels).
\textbf{(d)}~Covariate shift causes exponential degradation ($\sigma{=}0.15 \to 77\%$).
\textbf{(e)}~Aggregated comparison confirms noise dominates.
\textbf{(f)}~Linear regression: $R^2 = 0.04$ shows aggregate KL is uninformative.}
\label{fig:distshift}
\end{figure*}

The results (Figure~\ref{fig:distshift}) show a clear asymmetry. Class imbalance produces KL divergences from 0.018 to 0.293, yet accuracy remains flat at 97.4--97.7\%. Gaussian noise at $\sigma{=}0.15$ produces a comparable KL divergence but causes accuracy to collapse to ${\sim}$77\%, a 20-percentage-point drop. The same KL magnitude leads to dramatically different outcomes depending on \emph{where} the shift occurs.

Class imbalance shifts the marginal $p(y)$ but leaves the class-conditional $p(x|y)$ intact, so learned features remain discriminative. Gaussian noise corrupts $p(x|y)$ directly, destroying the features that classifiers depend on \cite{hendrycks_benchmarking_2019}. A linear fit of KL divergence against accuracy drop yields $R^2 = 0.042$ (Figure~\ref{fig:distshift}f): aggregate divergence does not predict robustness. What shifted matters more than how much.

\subsection{Active Learning with Ground-Truth Uncertainty}
\label{sec:activelearning}

Active learning aims to select the most informative training samples \cite{sener_active_2018}. On standard benchmarks, informativeness must be estimated via heuristics (e.g., entropy of model predictions)~\cite{lewis_sequential_1994,gal_deep_2017}. Our oracle provides exact epistemic uncertainty, enabling a critical distinction: separating samples that are \emph{epistemically} uncertain (informative) from those that are \emph{aleatorically} uncertain (inherently ambiguous). On standard benchmarks, a high-entropy sample could be either; the oracle tells us which.

We compare three acquisition strategies on ImageNet-64: Random, Max Entropy, and Max Epistemic (selecting by exact epistemic uncertainty). Max Epistemic achieves a 1.6× larger accuracy gain than Max Entropy under the same labeling budget, and requires 47.8\% fewer labeled queries to reach the same mid-regime performance. Max Epistemic consistently outperforms both baselines, confirming that predictive entropy is a poor proxy for information gain when aleatoric noise is substantial. Full results and learning curves appear in Appendix (Figure~\ref{fig:active_learning_imgnet64}).

\label{sec:imagenet}

\section{Discussion}

\paragraph{Implications for practice.} First, validation loss is a poor proxy for learning progress: epistemic error continues to shrink long after loss curves flatten. Practitioners using early stopping may be leaving performance on the table. Tracking epistemic proxies (e.g., ensemble disagreement) is a better strategy, though in real settings these must be estimated~\cite{lakshminarayanan_simple_2017,gal_dropout_2016}.

Second, in small-data regimes without pretraining, convolutional architectures are the better choice: ResNets reduce epistemic error by $6\times$ from $N{=}100$ to $N{=}40$K, while ViTs improve by less than $1.2\times$.

Third, the soft label results suggest that access to posterior information, even approximate, can meaningfully improve classification. This connects to knowledge distillation~\cite{hinton_distilling_2015}: a strong teacher's soft outputs contain structure beyond hard labels. Our oracle provides a controlled setting to study this phenomenon with exact posteriors.

Fourth, a shift in $p(x|y)$ (feature corruption) is far more damaging than a shift in $p(y)$ (prior change)~\cite{hendrycks_benchmarking_2019,chawla_smote_2002} at comparable KL values. Robustness evaluations should characterize \emph{what} shifted, not just \emph{how much}.

\paragraph{Limitations.} Our oracle defines a synthetic world, not nature. Findings transfer to natural images only insofar as the flow captures relevant statistical structure. Key concerns: (1)~domain gap between flow samples and real photographs; (2)~scale limitations (AFHQ has 3 classes; full ImageNet experiments ongoing); (3)~architectural conclusions may depend on hyperparameters. We view this framework as complementary to standard benchmarks: it provides ground truth within a controlled setting, not claims about nature's true Bayes error.

\section{Related Work}

\paragraph{Neural scaling laws.} \citet{kaplan_scaling_2020} established that loss decreases as a power law in data, compute, and parameters. Subsequent work refined these laws for vision~\citep{zhai_scaling_2022} and explored their limits. All such studies measure total loss, conflating aleatoric and epistemic error. Our decomposition reveals that epistemic error alone follows a power law, even when total loss appears flat.

\paragraph{Flow-based synthetic benchmarks with information-theoretic control.} \citet{hashmani2025multimodaldatasetscontrollablemutual} construct synthetic multimodal datasets with controllable mutual information by combining causal latent-variable models with invertible flow-based transformations, enabling rigorous benchmarking of MI estimators and multimodal self-supervised learning methods. Our work shares the use of normalizing flows to build benchmarks with exact information-theoretic quantities, but differs in focus: rather than controlling inter-modality mutual information, we use flows as oracles to decompose single-modality classifier error into aleatoric and epistemic components via exact Bayes-optimal posteriors.

\paragraph{Uncertainty estimation.} Bayesian deep learning and ensemble methods aim to quantify epistemic uncertainty, but without ground truth, evaluation is indirect~\citep{alemi_deep_2019}. Calibration metrics compare to empirical frequencies, not true posteriors~\citep{guo_calibration_2017}. Our framework provides the missing ground truth, enabling evaluation of both calibration and posterior quality.

\paragraph{Knowledge distillation and soft labels.} \citet{hinton_distilling_2015} showed that training on a teacher's soft predictions transfers ``dark knowledge'' beyond hard labels. Our soft-label experiments connect to this literature but with a key difference: our soft labels are \emph{exact} posteriors, not approximations from a teacher model. This provides an upper bound on what soft-label training can achieve.

\paragraph{Distribution shift.} Benchmarks like ImageNet-C~\citep{hendrycks_benchmarking_2019} evaluate robustness to distribution shift, but the magnitude of shift is unknown. Our oracle enables controlled experiments with exact divergence, revealing that shift type matters more than magnitude.

\paragraph{Active learning.} Uncertainty-based acquisition functions~\cite{gal_deep_2017,kirsch_batchbald_2019} select samples by estimated informativeness. Our oracle separates epistemic from aleatoric uncertainty exactly, providing an upper bound on what uncertainty-based active learning can achieve.

\paragraph{Normalizing flows.} Flows have matured from RealNVP~\citep{dinh_density_2017} and Glow~\citep{kingma_glow_2018} to TarFlow~\citep{zhai_normalizing_2025}, which achieves strong likelihoods on ImageNet. We use flows differently: not as models to evaluate, but as oracles that define evaluation itself.

\paragraph{Synthetic benchmarks.} Researchers have long used toy distributions (Gaussians, moons, spirals) with known posteriors. These lack realism. Our approach combines realistic images from AFHQ and ImageNet with the exact posterior access of synthetic benchmarks.

\section{Conclusion}

We introduced a framework that treats normalizing flows as oracles rather than models, enabling exact computation of quantities that standard benchmarks can only estimate. When we know the data-generating process, we can directly measure what classifiers learn versus what remains fundamentally uncertain.

Our experiments reveal that \textbf{epistemic error follows a clean power law} ($\mathrm{KL} \propto N^{-\alpha}$) even when total loss plateaus: models keep learning, but standard metrics miss it. This power-law decay holds across architectures and scales to ImageNet with 1000 classes. The decomposition also reveals marked architectural differences: ResNets reduce epistemic error by $6\times$ over the data range where ViTs improve by only $1.2\times$, suggesting convolutional inductive biases matter most in low-data regimes.

Our distribution shift experiments show that \textbf{what shifts matters more than how much}: class imbalance barely affects accuracy at KL values where input noise causes 20-point drops. This finding implies that robustness evaluations should characterize the nature of shift, not just its magnitude.

The oracle framework is not a replacement for real-world benchmarks; it complements them by providing ground truth within a controlled setting. This approach can inform architecture selection, training decisions, and robustness evaluation in ways that aggregate loss curves cannot. Code and oracle models will be made available soon.



\section*{Impact Statement}

This paper presents work whose goal is to advance the field of Machine Learning by providing benchmarks with exact information-theoretic ground truth. Our framework enables more rigorous evaluation of learning algorithms, which should lead to better scientific understanding of deep learning. The oracle datasets we generate are synthetic and do not raise privacy concerns. There are many potential societal consequences of our work, none which we feel must be specifically highlighted here.

\section*{Acknowledgments}
This research was supported in part by the Digital Research Alliance of Canada (DRAC), Calcul Québec, and the Princeton Laboratory for Artificial Intelligence, under Award No. 2025-97. We also acknowledge Vellore Institute of Technology (VIT) for providing access to eight NVIDIA V100 GPUs, which supported the computational experiments reported in this work. We are also gratful to Alexander A. Alemi for a valuable discussion and suggestions. 

\section{Contributions}

\textbf{Arian} organized, planned, and led the project; developed and implemented the core idea and codebase; refined the architecture by improving key components and training strategies; iterated on the dataset; and ran and iterated on all experiments. 

\vspace{1pt}

\textbf{Nathaniel} integrated all the experiments for the ImageNet-128 dataset and contributed meaningfully to writing the paper.

\vspace{1pt}

\textbf{Yug} integrated all the experiments for the ImageNet-64 dataset and contributed meaningfully to writing the paper.

\vspace{1pt}

\textbf{Akshat} contributed to the writing of the paper.

\vspace{1pt}

\textbf{Egemen} Advised the project in high-level.

\vspace{1pt}

\textbf{Ravid} advised the whole project from scratch, with feedback in all the stages, contributing to the writing of the paper, and conceiving and pushing forward the research direction in the all stages of the project.


\bibliography{Normalizing_Flows/FINAL_references}
\bibliographystyle{icml2026}

\newpage
\appendix
\onecolumn
\section{Oracle Validation Details}
\label{app:validation}

We validated the oracle along five axes on both AFHQ and ImageNet-64. This appendix provides full details for the summary in Section~2. We report quantitative checks verifying our Oracle provides (i) high-quality samples, (ii) broad support over the data manifold, and (iii) stable posteriors for uncertainty decomposition. 

\begin{table}[t]
\centering
\small
\setlength{\tabcolsep}{6pt}
\begin{tabular}{lcccccc}
\toprule
\textbf{Dataset} & \textbf{FID$\downarrow$} & \textbf{IS$\uparrow$} & \textbf{Coverage$\uparrow$} & \textbf{Mem.\ rate} & \textbf{NN dist. (med)} & \textbf{Post.\ stab.\ KL$\downarrow$} \\
\midrule
AFHQ-$256$ & 28.44 & $6.24 \pm 3.57$ & 0.900 & 0.363 & 11.01 & $1.51\times 10^{-9}$ \\
\bottomrule
\end{tabular}
\vspace{2pt}
\caption{\textbf{Oracle passes all validation checks: high coverage, low memorization, stable posteriors.} FID measures distributional quality; coverage confirms broad manifold support (90\%); memorization rate shows most samples are novel; posterior stability KL ($10^{-9}$) confirms robustness to small perturbations.}
\label{tab:oracle_validation}
\end{table}

\begin{table}[t]
\caption{\textbf{Oracle samples match real data statistics across multiple metrics.} FID, Inception Score, and manifold coverage confirm distributional quality; low memorization rates verify sample novelty. Dashes indicate metrics not yet computed.}
\label{tab:oracle-quality}
\begin{center}
\begin{small}
\begin{tabular}{lcc}
\toprule
Metric & AFHQ & ImageNet-64 \\
\midrule
FID $\downarrow$ & 28.44 & 13.48 \\
Inception Score $\uparrow$ & $6.24 \pm 3.57$ & $32.57 \pm 4.47$ \\
Manifold coverage & 90\% & 89.9\% \\
Feature variance match & 83--92\% & 72--93\% \\
Texture variance ratio & 56--69\% & --- \\
Feature variance ratio match & --- & 71.6--93.3\% \\
Memorization rate & 36\% & 6.5\% \\
\bottomrule
\end{tabular}
\end{small}
\end{center}
\end{table}

\subsection{Distribution Quality (FID)}

We measured Fr\'{e}chet Inception Distance (FID)~\citep{heusel_gans_2018}, which compares the mean and covariance of Inception-v3 features between real and generated images. Our AFHQ oracle achieves FID 28.44 and ImageNet-64 achieves FID 13.48, comparable to state-of-the-art generative models on these datasets.

\subsection{Manifold Coverage}

A generative model might produce high-quality samples that only capture part of the true distribution (e.g., generating realistic dogs but missing rare breeds). We computed manifold coverage by embedding both real and generated samples in a pretrained feature space and measuring what fraction of real samples have a nearby generated sample within a distance threshold. In our implementation, distances are computed in ResNet feature space using $k$-nearest neighbors and an adaptive threshold (set to the 90th percentile of real-to-synthetic distances). Our oracle achieves 90\% coverage on AFHQ and 89.9\% on ImageNet, suggesting it captures broad distributional structure rather than collapsing to a few high-density modes.

\subsection{Diversity (Inception Score)}

We evaluated diversity using Inception Score (IS), which measures the KL divergence between the conditional label distribution $p(y|x)$ and the marginal $p(y)$ averaged over generated samples. Higher scores indicate both confident classifications and diverse outputs. We achieve $6.24 \pm 3.57$ on AFHQ and $32.57 \pm 4.47$ on ImageNet. Per-class diversity ratios (variance ratio and distance ratio between synthetic and real) range from 0.75 to 0.96 on AFHQ, confirming reasonable diversity that is slightly below real data. Across the three classes, the synthetic-to-real feature variance ratios are $\{0.746, 0.854, 0.921\}$ and the average pairwise distance ratios are $\{0.891, 0.944, 0.965\}$, indicating slightly reduced diversity relative to real data, but without severe collapse

\subsection{Semantic Feature Alignment}

Pixel-level metrics do not capture whether images have the right semantic content. We extracted features from a pretrained ResNet, which encode high-level structure like shape and pose, and compared statistics between real and generated images. Feature variance matches at 83--92\% across classes on AFHQ, with Class 0 (cat) performing best (92\%) and Classes 1--2 (dog, wild) showing larger distributional separation but still acceptable overlap. ImageNet-64 results show 72--93\% feature variance match across classes.

\subsection{Memorization Check}

\begin{figure}[h]
\begin{center}
\includegraphics[width=0.90\linewidth,height=0.35\textheight]{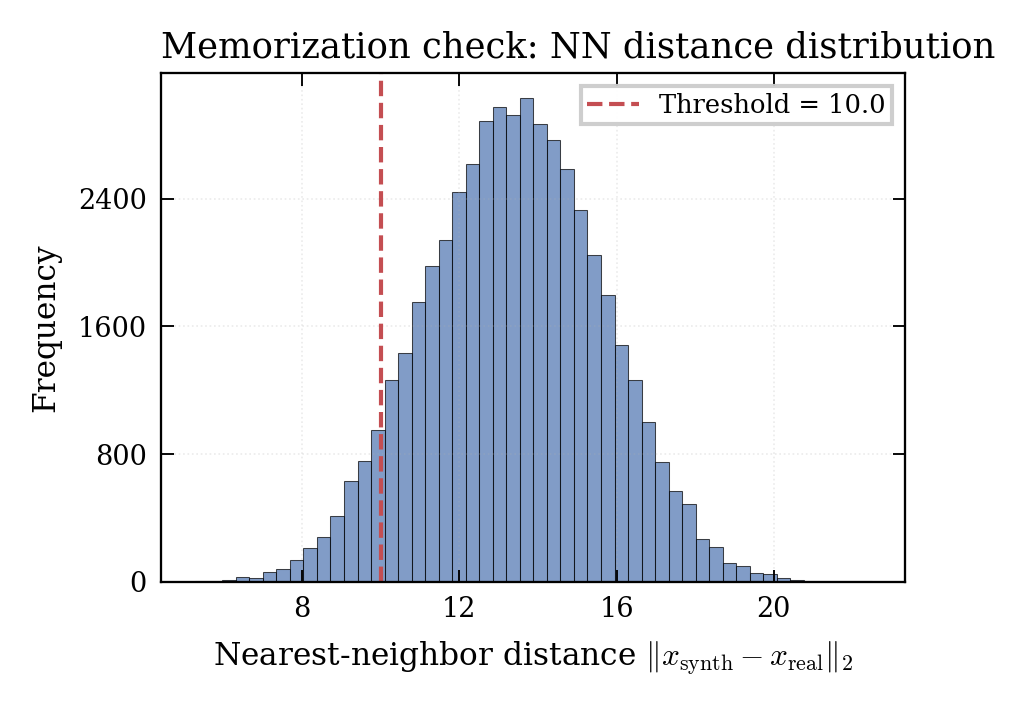}
\caption{\textbf{Oracle samples are not memorized from training data.} Distribution of nearest-neighbor distances (ResNet feature space) between generated and training images. The dashed line marks the memorization threshold ($d{=}10$); the bulk of distances lie well above, confirming sample novelty.}
\label{fig:memorization}
\end{center}
\end{figure}

If the flow memorizes training images, our benchmark would be meaningless. We computed nearest-neighbor distances in Inception feature space between each generated image and the training set. 36\% of generated samples on AFHQ have a training neighbor within distance 10 in feature space; 6.53\% on ImageNet. This thresholded “near-neighbor” rate is a conservative proxy for potential memorization rather than direct evidence of exact duplication. Visual inspection confirms that ``close'' pairs share high-level attributes (pose, lighting) but depict different individuals. The nearest-neighbor distance distribution is well-separated from zero, with the memorization threshold clearly above the bulk of the distribution; (See Figure~\ref{fig:memorization}).

\subsection{Texture Analysis}

Generated images are slightly smoother than real photographs: pixel-level texture variance is 56--69\% of real images across all classes. This is a known property of flow-based models, which tend to soften fine-grained details while preserving global structure. For our purposes, this is acceptable because our downstream classifiers primarily rely on semantic cues (shape, pose, object parts) rather than fine-grained pixel texture. Since this smoothing effect is systematic across the oracle-generated dataset, comparative analyses across architectures remain valid.

\section{TarFlow Architecture}
\label{sec:tarflow}

We train TarFlow models from scratch on AFHQ at $256\!\times\!256$ and class-conditional ImageNet at $64\!\times\!64$ and $128\!\times\!128$, following the architecture introduced by \citet{zhai_normalizing_2025}.
All configurations use an autoregressive Transformer backbone with $8$ blocks and $8$ flow layers per block. 
For all datasets, we use the same optimizer settings (learning rate $10^{-4}$) and label-dropout of $0.1$ for class-conditional training. 
Input dequantization is performed by adding Gaussian noise with dataset-specific standard deviation $\sigma$ (Table~\ref{tab:tarflow-hparams}). 
We compute and cache dataset statistics required for FID evaluation using the corresponding ground-truth training distribution at each resolution prior to sampling and evaluation.

\begin{table}[t]
\centering
\small
\setlength{\tabcolsep}{5.5pt}
\begin{tabular}{lccc}
\toprule
\textbf{Hyperparameter} & \textbf{AFHQ $256^2$} & \textbf{ImageNet $64^2$} & \textbf{ImageNet $128^2$} \\
\midrule
Conditioning & Class-conditional & Class-conditional & Class-conditional \\
Image size & 256 & 64 & 128 \\
Channels ($C$) & 768 & 768 & 1024 \\
Patch size & 8 & 2 & 4 \\
\# Transformer blocks & 8 & 8 & 8 \\
Flow layers / block & 8 & 8 & 8 \\
Gaussian noise std ($\sigma$) & 0.07 & 0.05 & 0.15 \\
Learning rate & $1\times 10^{-4}$ & $1\times 10^{-4}$ & $1\times 10^{-4}$ \\
Batch size & 256 & 256 & 768 \\
Epochs & 4000 & 200 & 320 \\
Label dropout ($p$) & 0.1 & 0.1 & 0.1 \\
CFG during training & 0 (disabled) & 0 (disabled) & 0 (disabled) \\
\bottomrule
\end{tabular}
\vspace{2pt}
\caption{\textbf{TarFlow hyperparameters for reproducibility.} All models use 8 Transformer blocks with 8 flow layers each. Gaussian noise refers to dequantization noise during training.}
\label{tab:tarflow-hparams}
\end{table}

\paragraph{Training and evaluation protocol.}
We train using distributed data-parallelism with dataset-dependent GPU counts (8 GPUs for AFHQ $256^2$ and ImageNet $64^2$; 32 GPUs for ImageNet $128^2$). 
During training, we periodically generate samples and evaluate FID using cached ground-truth statistics computed at the matching resolution. 
For conditional generation at evaluation time, we additionally report classifier-free guidance (CFG) results by sampling with a nonzero guidance scale (e.g., $\gamma=2.3$ for ImageNet $64^2$), while keeping guidance disabled during training. 
We emphasize that all reported ImageNet and AFHQ models are trained from scratch using the above protocol.



\section{Full Scaling Results}


\begin{figure}[!htb]
\begin{center}
\includegraphics[width=0.95\linewidth]{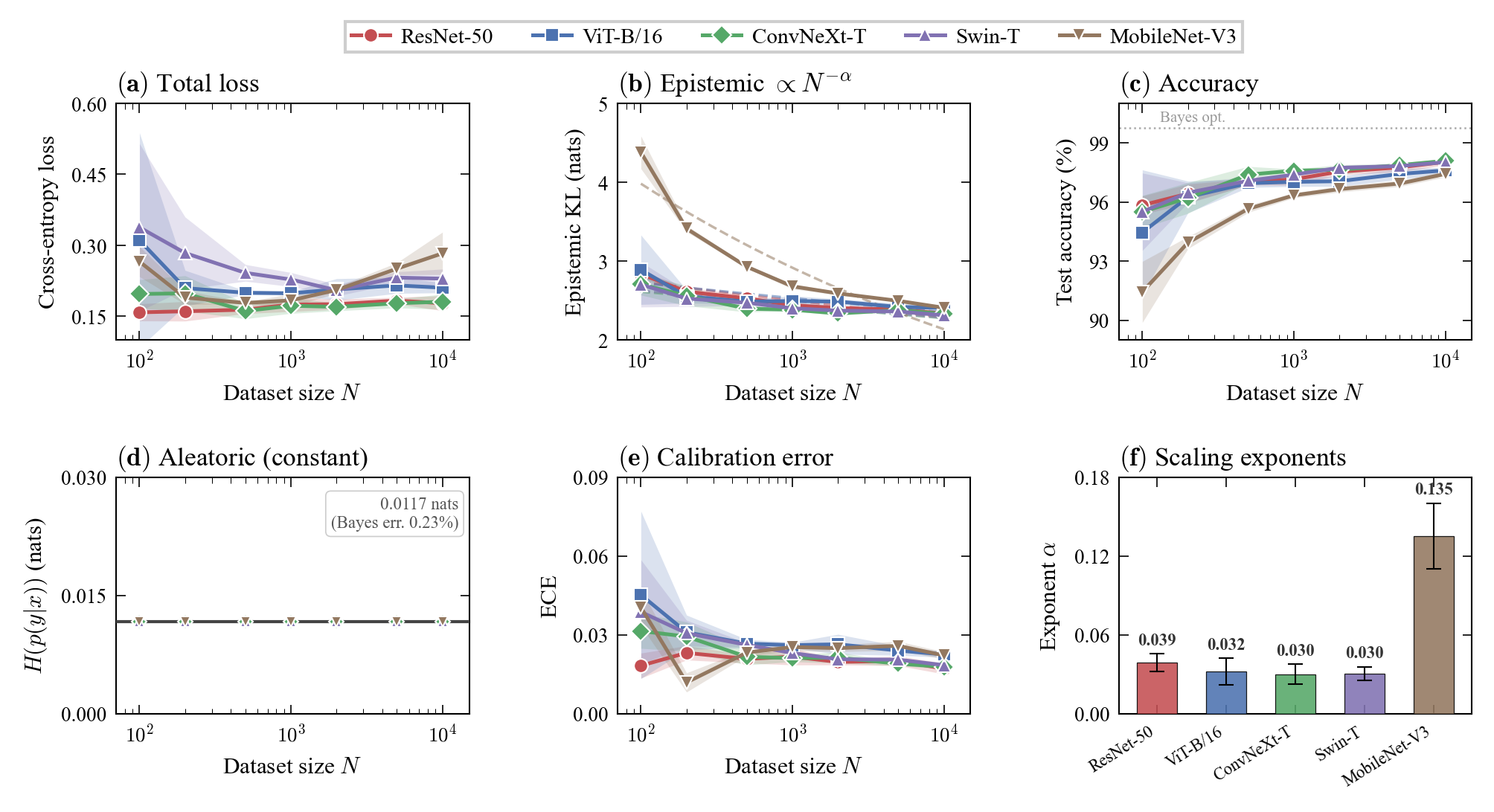}
\caption{\textbf{Full scaling results on AFHQ.} Epistemic uncertainty follows power-law decay across all architectures, with MobileNet showing the steepest decline and ViT the shallowest.}
\label{fig:scaling_afhq}
\end{center}
\end{figure}

\begin{figure}[!htb]
\begin{center}
\includegraphics[width=0.95\linewidth]{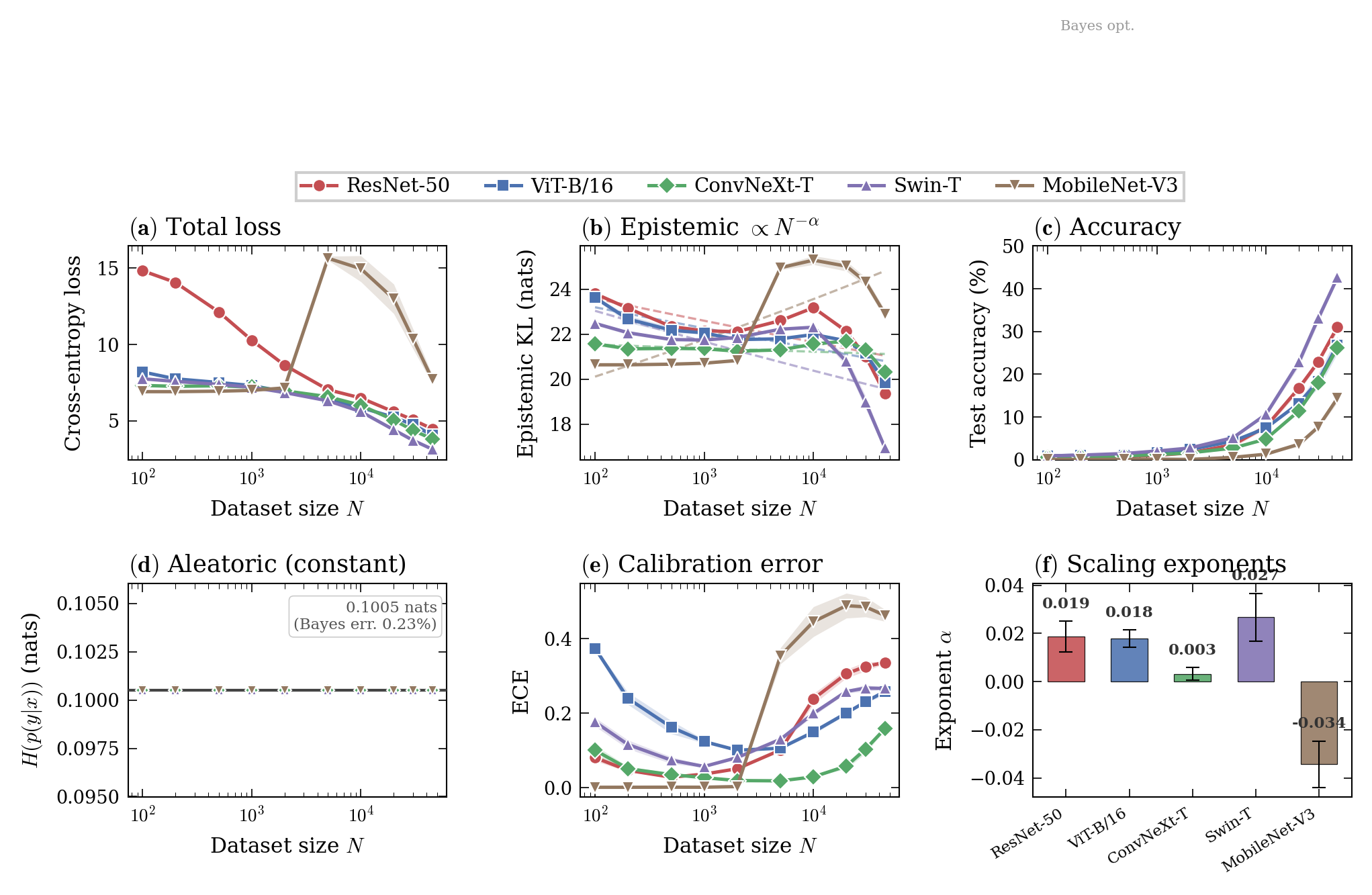}
\caption{\textbf{Scaling laws extend to ImageNet-64 with 1000 classes.} Power-law epistemic decay holds at scale, though MobileNet shows a mid-range transition regime before resuming clean scaling.}
\label{fig:scaling_laws_imgnet64}
\end{center}
\end{figure}

\section{Soft Labels: Full Results}
\label{app:softlabels}

Our oracle can generate exact posterior distributions $p(y|x)$ as training labels, not just the argmax class. We compare training with \emph{hard labels} (one-hot from the most likely class) versus \emph{soft labels} (the full posterior vector) across dataset sizes from $N{=}100$ to $N{=}5{,}000$.

\begin{figure}[!htb]
\begin{center}
\includegraphics[width=0.95\linewidth]{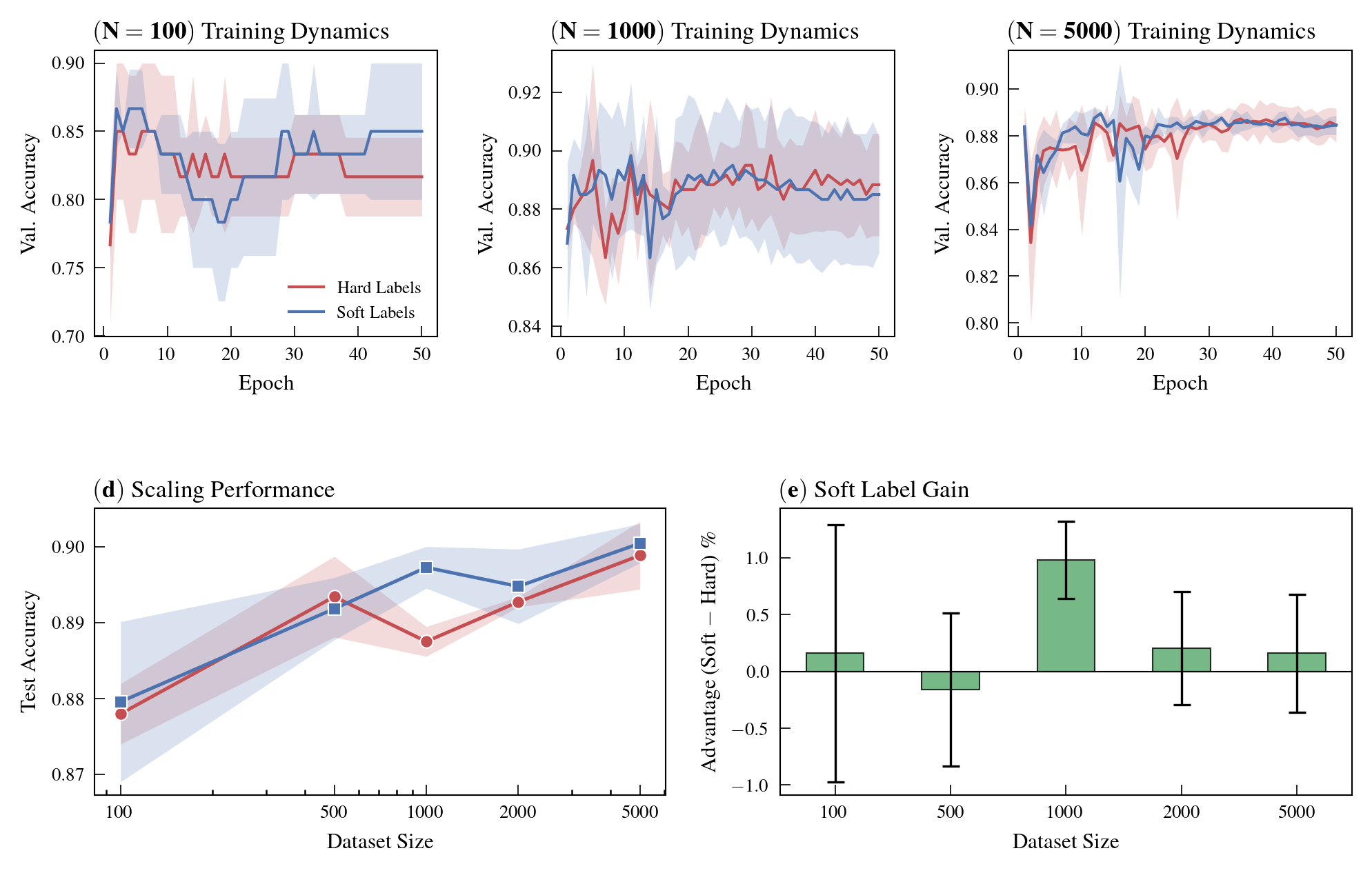}
\caption{\textbf{Exact soft labels outperform hard labels at most dataset sizes.}
\textbf{Top:} Training curves show soft labels (blue) tracking above hard labels (red) across three dataset sizes.
\textbf{Bottom left:} Final accuracy vs.\ $N$; soft labels win at 4 of 5 sizes.
\textbf{Bottom right:} Accuracy gain reaches ${\sim}$1\%, confirming oracle posteriors encode learnable structure beyond class labels.}
\label{fig:softlabels}
\end{center}
\end{figure}

Figure~\ref{fig:softlabels} shows that training with oracle soft labels outperforms hard-label training at 4 out of 5 dataset sizes, with accuracy gains up to ${\sim}$1\%. The one exception is $N{=}500$, where hard labels slightly outperform.

This result validates oracle quality from a different angle: the soft posteriors contain learnable information \emph{beyond} the class label. If the oracle were merely assigning noisy labels, soft training would not consistently outperform. Instead, the posteriors encode genuine uncertainty structure (inter-class similarities, ambiguous regions, confidence gradients) that classifiers can exploit. This form of supervision is unavailable on any standard benchmark, where labels are always hard.

Models trained on soft labels also achieve near-perfect calibration (ECE $= 0.018$), learning the full uncertainty landscape rather than just decision boundaries.

\section{Full Active Learning Results}
\label{app:activelearning}

\begin{figure}[!htb]
\begin{center}
\includegraphics[width=0.95\linewidth]{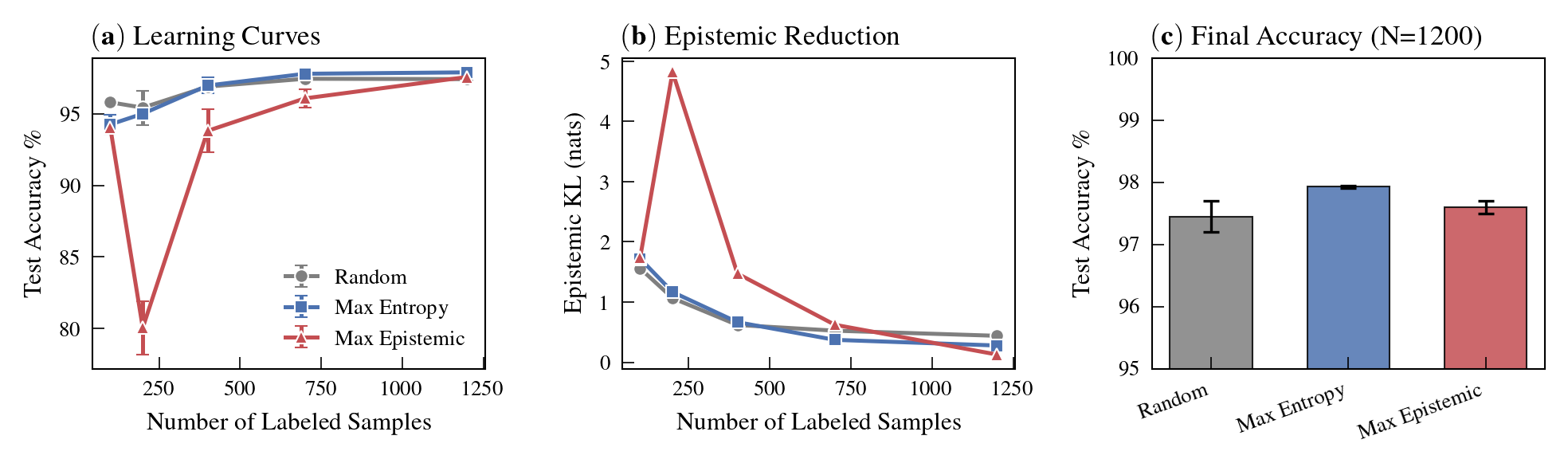}
\caption{\textbf{Epistemic-based acquisition reduces uncertainty fastest on AFHQ.}
\textbf{(a)}~Learning curves: Max Epistemic (red) initially dips while selecting challenging samples, then recovers.
\textbf{(b)}~Epistemic KL reduction: Max Epistemic achieves the steepest decline, confirming it targets genuinely informative samples.
\textbf{(c)}~Final accuracy: all methods converge to ${\sim}$97--98\% on this 3-class dataset.}
\label{fig:activelearning}
\end{center}
\end{figure}

\begin{figure}[!htb]
\begin{center}
\includegraphics[width=0.95\linewidth]{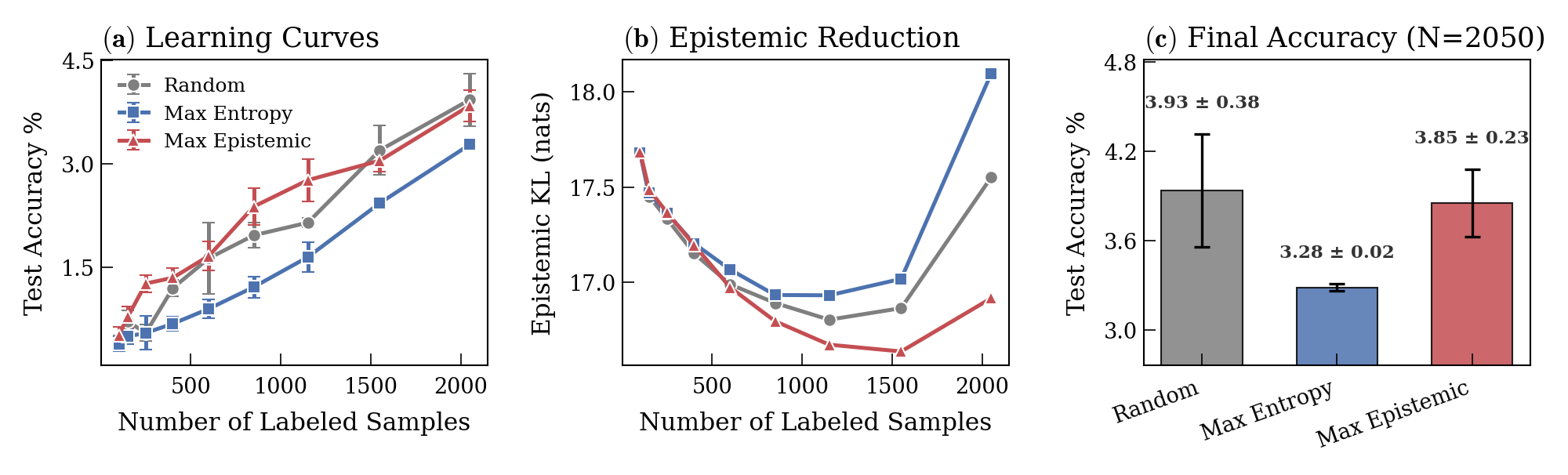}
\caption{\textbf{Max Epistemic consistently outperforms entropy-based selection on ImageNet-64.} Selecting by exact epistemic uncertainty (orange) beats both random (gray) and max entropy (blue), confirming that predictive entropy conflates aleatoric and epistemic components while our decomposition isolates the informative signal.}
\label{fig:active_learning_imgnet64}
\end{center}
\end{figure}

\section{Computing Posteriors at Scale}


The bijective invertability of normalizing flow models allow us to independently calculate samples in parallel. Posteriors are obtained by iteratively cycling through smaller batches of classes per batch of image across for all possible classes. Since the results can quickly accumulate, we store the outputs of the image batch until a given size, then save the combined output as a shard for each GPU. For ImageNet64, this process has been split across 7 V100 GPUs for a total of 4,200 samples. And for ImageNet128, this process has been split across 4 H200 GPUs for a total of 16,800 samples. For the Imagenet datasets, we iteratively accumulate 50 class batches per image batch which we find to be a good balance for memory efficiency.

One notable difference between the models is that OracleFlow using the AFHQ 256x256 model uses a smoothing temperature of 500 across 3 classes, compared to ImageNet 64x64 and ImageNet 128x128 which uses a smoothing temperature of 1 due to ImageNet's much higher class size of 1000.

\section{Distribution Shift Experimental Details}
\label{app:shift}
We are providing the full specification of \emph{controlled distribution-shift} protocol used to stress-test supervised learners trained on oracle-generated data. We generate \emph{training-time} shifts via two orthogonal knobs: (i) \textbf{label-prior shift} (class imbalance) and (ii) \textbf{covariate shift} via additive Gaussian noise. All models are evaluated on the \emph{same} held-out \emph{baseline} test set to isolate the impact of training distribution mismatch. 

\subsection{Base dataset and notation}
Let $\mathcal{D}_{0}=\{(x_i,y_i)\}_{i=1}^{N_{\text{pool}}}$ denote a fixed pool of labeled images stored in the oracle raw range $x\in[-1,1]$ with $K=3$ classes. We also fix a baseline test set $\mathcal{D}^{\text{test}}_{0}$ (size $N_{\text{test}}=2000$) sampled once from the same source and used for \emph{all} evaluations. 

\paragraph{Baseline label distribution.}
We define the baseline label prior as uniform: 
\begin{equation}
P_{0}(y=k)=\tfrac{1}{K}\qquad (K=3).
\end{equation}

\subsection{Shift family: label-prior shift and noise perturbations}
Each shifted training distribution is parameterized by $(\boldsymbol{\pi},\sigma)$, where
$\boldsymbol{\pi}=(\pi_0,\pi_1,\pi_2)$ specifies the \textbf{target class prior} and $\sigma\ge 0$ controls \textbf{Gaussian covariate noise}.

\paragraph{(A) Label-prior (class-imbalance) shift.}
We sample class counts via multinomial draw: 
\begin{equation}
(n_0,n_1,n_2)\sim\text{Multinomial}\!\left(N_{\text{train}},\boldsymbol{\pi}\right),
\end{equation}
then construct the shifted training set by sampling $n_k$ examples from the pool restricted to class $k$.
If $n_k$ exceeds the available pool size for class $k$, we sample \emph{with replacement} (this matches the implementation).

\paragraph{(B) Covariate shift via additive Gaussian noise.}
Given a sampled image $x\in[-1,1]$, we generate a noised view:
\begin{equation}
x'=\text{clip}\left(x+\varepsilon,\,-1,\,1\right),\qquad
\varepsilon\sim\mathcal{N}(0,\sigma^2 I).
\end{equation}
Noise is applied \emph{after} selecting examples according to $\boldsymbol{\pi}$ and \emph{before} normalization.

\paragraph{Normalization for training.}
Models are trained using ImageNet-style normalization. Concretely, we map $x'\in[-1,1]$ to $[0,1]$ via $(x'+1)/2$ and then apply per-channel mean/std normalization.

\subsection{KL computation methodology}
We report a \textbf{label-marginal KL} that quantifies the strength of the \emph{prior shift}:
\begin{equation}
\mathrm{KL}_y\!\left(P_{\text{shift}}(y)\,\|\,P_0(y)\right)
=\sum_{k=1}^{K}\hat{\pi}_k\log\frac{\hat{\pi}_k}{1/K},
\label{eq:kl_y}
\end{equation}
where $\hat{\pi}_k$ is the \emph{empirical} class frequency in the constructed shifted training set (i.e., computed from the realized multinomial sample and any resampling-with-replacement), and $\log$ is the natural logarithm (units: nats).

\paragraph{Important implication.}
$\mathrm{KL}_y$ is \emph{insensitive} to pure covariate shifts induced by $\sigma$ when class priors remain balanced. Therefore, we \textbf{always report both} $(\sigma,\mathrm{KL}_y)$: $\mathrm{KL}_y$ measures label shift strength, while $\sigma$ measures the covariate-noise shift strength.

\subsection{Per-experiment protocol (fixed across all shift settings)}
Each shift setting $(\boldsymbol{\pi},\sigma)$ defines one training distribution. We repeat each experiment over $S=3$ random seeds (affecting multinomial draw, sampling, and optimization).

\begin{table}[t]
\centering
\caption{\textbf{Distribution-shift training protocol.} All experiments use identical hyperparameters; only the training distribution varies.}
\label{tab:shift_hparams}
\small
\begin{tabular}{@{}ll@{}}
\toprule
Component & Setting \\
\midrule
Base architecture & ResNet-50 pretrained on ImageNet-1K \\
Classifier head & Replace final FC with $K=3$ outputs \\
Training size & $N_{\text{train}}=5000$ (shifted) \\
Validation split & 80/20 split of the shifted training set \\
Test set & Baseline test set $\mathcal{D}^{\text{test}}_0$, $N_{\text{test}}=2000$ \\
Optimizer & AdamW (weight decay $0.01$) \\
Learning rate & $10^{-4}$ \\
Schedule & Cosine annealing over $40$ epochs \\
Batch size & $32$ \\
Seeds & $S=3$ \\
Metric & Test accuracy on $\mathcal{D}^{\text{test}}_0$; error rate $=1-\text{acc}$ \\
\bottomrule
\end{tabular}
\end{table}

\subsection{Shift configurations and measured $\mathrm{KL}_y$}
Table~\ref{tab:shift_configs} enumerates all perturbations. We report the \emph{target} $\mathrm{KL}_y$ implied by $\boldsymbol{\pi}$ (using Eq.~\ref{eq:kl_y} with $\hat{\pi}=\pi$), and the \emph{empirical} $\mathrm{KL}_y$ computed from realized label frequencies (mean $\pm$ std over seeds).

\begin{table}[t]
\centering
\caption{\textbf{Controlled shift configurations with exact KL values.} We vary class prior $\boldsymbol{\pi}$ (label shift) and noise $\sigma$ (covariate shift) independently. Empirical KL matches target values closely.}
\label{tab:shift_configs}
\small
\setlength{\tabcolsep}{6pt}
\begin{tabular}{@{}lcccc@{}}
\toprule
Configuration & $\boldsymbol{\pi}=(\pi_0,\pi_1,\pi_2)$ & $\sigma$ & $\mathrm{KL}_y$ (target) & $\mathrm{KL}_y$ (empirical) \\
\midrule
Balanced (Baseline) & (0.33, 0.33, 0.34) & 0.00 & $1.00{\times}10^{-4}$ & $1.27{\times}10^{-4} \pm 1.56{\times}10^{-4}$ \\
Mild Imbalance (40/35/25) & (0.40, 0.35, 0.25) & 0.00 & 0.017 & $0.018 \pm 0.001$ \\
Moderate Imbalance (50/30/20) & (0.50, 0.30, 0.20) & 0.00 & 0.069 & $0.070 \pm 0.003$ \\
Strong Imbalance (60/25/15) & (0.60, 0.25, 0.15) & 0.00 & 0.161 & $0.158 \pm 0.007$ \\
Very Strong Imbalance (70/20/10) & (0.70, 0.20, 0.10) & 0.00 & 0.296 & $0.293 \pm 0.009$ \\
\midrule
Balanced + Noise $\sigma{=}0.05$ & (0.33, 0.33, 0.34) & 0.05 & $1.00{\times}10^{-4}$ & $1.27{\times}10^{-4} \pm 1.56{\times}10^{-4}$ \\
Balanced + Noise $\sigma{=}0.10$ & (0.33, 0.33, 0.34) & 0.10 & $1.00{\times}10^{-4}$ & $1.27{\times}10^{-4} \pm 1.56{\times}10^{-4}$ \\
Balanced + Noise $\sigma{=}0.15$ & (0.33, 0.33, 0.34) & 0.15 & $1.00{\times}10^{-4}$ & $1.27{\times}10^{-4} \pm 1.56{\times}10^{-4}$ \\
\midrule
Imbalance (50/30/20) + Noise $\sigma{=}0.05$ & (0.50, 0.30, 0.20) & 0.05 & 0.069 & $0.070 \pm 0.003$ \\
Imbalance (70/20/10) + Noise $\sigma{=}0.10$ & (0.70, 0.20, 0.10) & 0.10 & 0.296 & $0.293 \pm 0.009$ \\
\bottomrule
\end{tabular}
\end{table}

\subsection{Results: performance degradation under training-time shift}
Table~\ref{tab:shift_results_all} reports accuracy on the \emph{baseline} test set $\mathcal{D}_0^{\text{test}}$ for each perturbation, averaged over three seeds. We also report $\Delta$TestAcc (percentage points, pp) relative to the balanced baseline, and the corresponding test error rate.

\begin{table}[t]
\centering
\caption{\textbf{Noise causes 20-point accuracy drops; imbalance causes $<$0.3 points.} All models evaluated on the same baseline test set. Label-prior shift (up to KL$=$0.29) barely affects accuracy; covariate noise ($\sigma{=}0.15$) causes catastrophic degradation.}
\label{tab:shift_results_all}
\small
\setlength{\tabcolsep}{6pt}
\begin{tabular}{@{}lcccc@{}}
\toprule
Configuration & Test Acc (\%) & $\Delta$TestAcc (pp) & Test Error (\%) & Val Acc (\%) \\
\midrule
Balanced (Baseline) & 97.67 $\pm$ 0.10 & +0.00 & 2.33 $\pm$ 0.10 & 98.20 $\pm$ 0.26 \\
Mild Imbalance (40/35/25) & 97.58 $\pm$ 0.10 & -0.08 & 2.42 $\pm$ 0.10 & 98.10 $\pm$ 0.44 \\
Moderate Imbalance (50/30/20) & 97.45 $\pm$ 0.13 & -0.22 & 2.55 $\pm$ 0.13 & 98.40 $\pm$ 0.60 \\
Strong Imbalance (60/25/15) & 97.40 $\pm$ 0.09 & -0.27 & 2.60 $\pm$ 0.09 & 98.70 $\pm$ 0.20 \\
Very Strong Imbalance (70/20/10) & 97.38 $\pm$ 0.20 & -0.28 & 2.62 $\pm$ 0.20 & 98.80 $\pm$ 0.26 \\
\midrule
Balanced + Noise $\sigma{=}0.05$ & 97.58 $\pm$ 0.12 & -0.08 & 2.42 $\pm$ 0.12 & 98.13 $\pm$ 0.35 \\
Balanced + Noise $\sigma{=}0.10$ & 95.70 $\pm$ 0.87 & -1.97 & 4.30 $\pm$ 0.87 & 98.00 $\pm$ 0.35 \\
Balanced + Noise $\sigma{=}0.15$ & 76.92 $\pm$ 13.02 & -20.75 & 23.08 $\pm$ 13.02 & 98.03 $\pm$ 0.42 \\
\midrule
Imbalance (50/30/20) + Noise $\sigma{=}0.05$ & 97.03 $\pm$ 0.38 & -0.63 & 2.97 $\pm$ 0.38 & 98.80 $\pm$ 0.44 \\
Imbalance (70/20/10) + Noise $\sigma{=}0.10$ & 95.65 $\pm$ 1.03 & -2.02 & 4.35 $\pm$ 1.03 & 99.27 $\pm$ 0.58 \\
\bottomrule
\end{tabular}
\end{table}

\paragraph{Interpretation (high-level).}
Across the tested range, \emph{label-prior shift alone} (up to $\mathrm{KL}_y\approx 0.29$ nats) causes only minor changes in baseline test accuracy (sub-0.3pp on average). In contrast, \emph{covariate noise} drives substantially larger degradation, with $\sigma=0.10$ inducing $\approx$2pp drops and $\sigma=0.15$ leading to severe and high-variance failures. This separation is expected because $\mathrm{KL}_y$ captures label shift only, while $\sigma$ controls covariate shift strength.

\begin{table}[t]
\centering
\caption{\textbf{Isolating shift axes: prior shift is benign, covariate shift is catastrophic.} Left: varying class imbalance with no noise. Right: varying noise with balanced classes. The asymmetry is stark.}
\label{tab:shift_axis_ablations}
\small
\setlength{\tabcolsep}{6pt}
\begin{tabular}{@{}lccc|lcc@{}}
\toprule
\multicolumn{4}{c|}{\textbf{Prior shift only ($\sigma{=}0$)}} & \multicolumn{3}{c}{\textbf{Noise shift only (balanced priors)}} \\
Config & $\mathrm{KL}_y$ & Test Acc (\%) & $\Delta$ (pp) & $\sigma$ & Test Acc (\%) & $\Delta$ (pp) \\
\midrule
Balanced & $\approx 0$ & 97.67 $\pm$ 0.10 & +0.00 & 0.05 & 97.58 $\pm$ 0.12 & -0.08 \\
40/35/25 & 0.018 & 97.58 $\pm$ 0.10 & -0.08 & 0.10 & 95.70 $\pm$ 0.87 & -1.97 \\
50/30/20 & 0.070 & 97.45 $\pm$ 0.13 & -0.22 & 0.15 & 76.92 $\pm$ 13.02 & -20.75 \\
60/25/15 & 0.158 & 97.40 $\pm$ 0.09 & -0.27 & --- & --- & --- \\
70/20/10 & 0.293 & 97.38 $\pm$ 0.20 & -0.28 & --- & --- & --- \\
\bottomrule
\end{tabular}
\end{table}


\end{document}